\documentclass[runningheads]{llncs}
\usepackage[T1]{fontenc}
\usepackage{color}

\usepackage{graphicx}
\usepackage{amssymb}

\begin{document}
\title{OCR is All you need: Importing Multi-Modality into Image-based Defect Detection System\thanks{This study was supported in part by ASE corporation; partly by the National Science and Technology Council (NSTC), Taiwan, under grants MOST 112-2221-E-006 -157 -MY3 and 111-2221-E-006 -210; partly by the Higher Education Sprout Project of Ministry of Education (MOE) to the Headquarters of University Advancement at National Cheng Kung University (NCKU); partly by High Education Humanities and Social Science Subjects Benchmark Project, Minstry of Education, Taiwan}}

%
\titlerunning{OCR is All you need}
%
\author{Chih-Chung Hsu\inst{1*},
Chia-Ming Lee\inst{1},
Chun-Hung Sun\inst{2},
Kuang-Ming Wu\inst{2}
} 
\institute{Institute of Data Science, National Cheng Kung University\inst{1} \\
      Corporate R$\&$D, Advanced Semiconductor Engineering Group\inst{2}\\
    \email{\inst{*}cchsu@gs.ncku.edu.tw}
}



\maketitle              

\begin{abstract}
Automatic optical inspection (AOI) plays a pivotal role in the manufacturing process, predominantly leveraging high-resolution imaging instruments for scanning purposes. It detects anomalies by analyzing image textures or patterns, making it an essential tool in industrial manufacturing and quality control.
Despite its importance, the deployment of models for AOI often faces challenges. These include limited sample sizes, which hinder effective feature learning, variations among source domains, and sensitivities to changes in lighting and camera positions during imaging. These factors collectively compromise the accuracy of model predictions.
Traditional AOI often fails to capitalize on the rich mechanism-parameter information from machines or inside images, including statistical parameters, which typically benefit AOI classification. 
To address this, we introduce an external modality-guided data mining framework, primarily rooted in optical character recognition (OCR), to extract statistical features from images as a second modality to enhance performance, termed OANet (Ocr-Aoi-Net). A key aspect of our approach is the alignment of external modality features, extracted using a single modality-aware model, with image features encoded by a convolutional neural network. This synergy enables a more refined fusion of semantic representations from different modalities. We further introduce feature refinement and a gating function in our OANet to optimize the combination of these features, enhancing inference and decision-making capabilities.
Experimental outcomes show that our methodology considerably boosts the recall rate of the defect detection model and maintains high robustness even in challenging scenarios.

\end{abstract}
\section{Introduction}

In the realm of industrial manufacturing, defect detection is paramount, serving a dual purpose: enhancing product quality and efficiency and indirectly curbing production costs by reducing the prevalence of both false negatives and positives. At its core, defect detection leverages a myriad of analytical instruments to discern features rooted in the physical properties of diverse products. Once these features are harvested, models trained on this data are then deployed for defect detection or classification.

Among the plethora of defect detection strategies, Automatic Optical Inspection (AOI) emerges as a predominant choice, especially in the industrial domain. This preference is not just due to its efficacy, but also because it incurs lower instrumentation costs compared to its analytical counterparts. AOI harnesses the power of high-resolution imaging tools to inspect products throughout the manufacturing phase. It taps into computer vision methodologies to spot defects, often focusing on attributes like shape, size, color, and product positioning. By juxtaposing the visual attributes of standard samples with defective ones, AOI can efficiently pinpoint anomalies. The convolutional neural network, in particular, has gained recognition for its prowess in extracting image features pivotal for defect inspection. However, the AOI methodology isn't without its limitations. For instance, burgeoning internal variability or an expanded dataset category spectrum can induce significant image feature fluctuations, complicating the defect detection process. Additionally, factors like unpredictable lighting conditions and even minuscule camera shifts can undercut the reliability of defect detection models. Even more challenges like sparse training data and sample imbalances can yield performance that falls short of expectations.

The dawn of multimodal learning has ushered in a new era where features from disparate modalities can be synergized, amplifying model performance. This integration has catalyzed advancements in domains like computer vision and natural language processing. However, training multimodal models often demands an extensive repository of high-quality, human-annotated image-text pairings, making the deployment of such models in AOI and defect detection less straightforward. Prevailing research has primarily zeroed in on fusing features from various modalities to sculpt models adept at multitasking across multiple modalities. However, the multimodal learning framework strictly requires multi-modality data as the input, which often becomes difficult for the AOI field since the machines usually do not allow the intrinsic data to be accessible. We judiciously challenge this issue. Since some statistical information could be printed in the AOI images in most cases, it could benefit the performance and robustness once the statistical information inside images is available.

In response to the challenges outlined above, we harness Optical Character Recognition (OCR) to morph external modality data within images into formats amenable to training single-modality-aware multimodal models. A focal point of our methodology is feature alignment prior to fusion, ensuring a harmonious amalgamation of features derived from different modalities. We also incorporate a gating function and feature refinement to maximize the potential of single modality features, especially crucial when another modal feature is compromised due to OCR malfunctions or data transmission distortions. Our experimental evaluations, utilizing datasets from ASE Corporation, attest to the efficacy of our approach.

\begin{itemize}
    \item S\textbf{ingle-Modality-Aware Multimodal Learning in AOI}: We challenge the conventional multimodal learning paradigm which often necessitates data from multiple modalities as input. Recognizing that machines in the AOI domain often restrict access to intrinsic data, we propose a novel approach that exploits statistical information typically embedded within AOI images. This insight allows us to enhance the performance and robustness of defect detection without relying on traditional multi-modality inputs.

\item \textbf{OCR-Driven Feature Extraction}: We introduce a pioneering method that employs Optical Character Recognition (OCR) to extract external modality data from images. This approach effectively converts such data into a format suitable for training single-modality-aware multimodal models, addressing the challenge of inaccessible intrinsic data in the AOI domain.

\item \textbf{Enhanced Feature Fusion Mechanism}: We present an advanced feature alignment process that is executed before fusion, ensuring optimal integration of features from distinct modalities. Accompanying this, we design a gating function and a feature refinement strategy. These are particularly critical when one modal feature might be compromised, either due to OCR errors or disruptions in data transmission. The effectiveness of our methods is empirically validated on the ASE Corporation dataset.
\end{itemize}





\section{Related Work}

\subsection{Advanced Defect Detection}

Defect detection has witnessed remarkable enhancements with the integration of deep and machine learning algorithms. These algorithms stand out for their proficiency in intricate feature extraction, managing vast data, pinpointing anomalies, and enabling real-time detection, thereby transcending the constraints of conventional methodologies. The realm of computer vision-based defect detection has particularly burgeoned, owing to the evolution of models adept at pixel-level classification.

Akcay et al. \cite{Anomalib} offer a suite of ready-to-deploy image-based anomaly and defect detection algorithms. Tabernik et al. \cite{Surface-Defect Detection} employ a semantic segmentation network for surface-defect detection, paving the way for fine-grained defect identification. Lopes et al. \cite{Auto-classifier} amalgamate the AutoML approach with various CNN-based models, aiming to bolster the robustness of the detection system.

\subsection{Multi-Modality Learning}

While traditional deep learning typically relies on single-modality data, recent advancements, such as CLIP \cite{CLIP}, have pioneered the integration of multiple modalities. Despite significant progress, challenges persist, including variations in dimensionality and semantic content across modalities, high costs of human annotation for paired data creation, and substantial computational requirements.

Pioneering works, such as those by Li et al. \cite{ALBEF}, have been instrumental in addressing these challenges. They have successfully integrated multi-modal contrastive learning with fusion learning, thus resolving semantic inconsistencies across different modalities. Their other contributions, including \cite{BLIP,BLIP-2,InstructBLIP2}, highlight the effectiveness of pre-training in reducing computational demands during the fine-tuning of multi-modal models.

Our work builds upon these advancements, yet diverges by focusing on generating new modalities from a single source. We explore the uncharted territory of harnessing statistical data from AOI images, a novel approach that promises to significantly advance the field of defect detection.

\subsection{Optical Character Recognition}

OCR technology, foundational in image processing and pattern recognition, is designed to convert printed or handwritten text into machine-readable data. While traditional OCR systems have their advantages, they face significant challenges, including the need for high-quality images, dependence on specific fonts and layouts, difficulties with complex backgrounds, and limitations in processing multilingual texts.

Advancements in deep learning have revolutionized OCR by dividing the process into distinct stages of text detection and recognition. Segmentation-based methods, such as those detailed in \cite{CTPN,SEGLINK,DBNET}, have been recognized for their precision in detecting text shape and position, particularly in complex layouts. Furthermore, the CNN-LSTM architecture \cite{CNN-LSTM,CNN-LSTM2} has been acclaimed for its effective balance between speed and accuracy, enhancing contextual understanding of the text.

In our research, we utilize OCR not only for its traditional role in text extraction but as a means to extract features from a single modality, laying the groundwork for our OANet. This innovative approach is designed to circumvent the typical challenges associated with OCR, particularly when dealing with compromised modal data.

\section{Methodology}

\subsection{Overview}
\begin{figure}
\includegraphics[width=\textwidth]{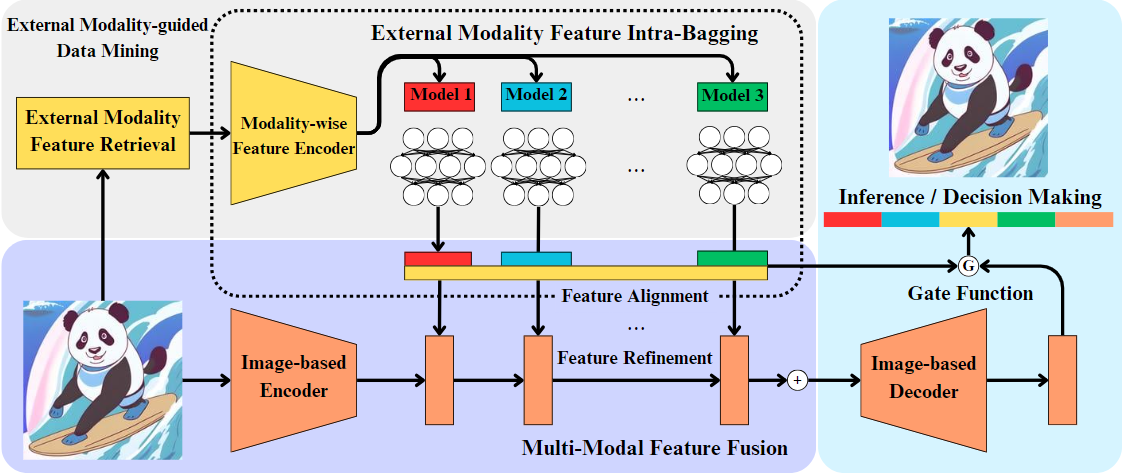}
\caption{The overall architecture of the proposed framework. It aims to explore external modality features to comprehensively support learning better semantic representations and decision-making. Several single modality features are extracted and fed into corresponding models, followed by feature alignment and refinement to enrich the feature space. Finally, we implement a gate function to aggregate all features and adaptively adjust the weights of different models or features, resulting in more robust and stable prediction results.}
 \label{framework2.png}
\end{figure}

This section delineates our novel method, designed to discern defects or anomalies within 2D images. A salient aspect of our approach lies in harnessing Optical Character Recognition (OCR) techniques to mine supplemental numerical or textual data from the images. Depending on the modality of this auxiliary information, we deploy appropriate sub-models for training. Post feature alignment, a feature refinement is executed on the attributes from the image encoder. This step amplifies the diversity and adaptability of the feature landscape. Upon decoding, predictions pertaining to the original images emerge. The conclusive inference draws upon the outputs of both the sub-models and the primary image model. The model's architecture is graphically depicted in Figure \ref{framework2.png}.

A unique strength of our method is its ability to assimilate information spanning diverse single modalities. The semantic richness harvested from these individual modalities augments the encoded image features, enhancing their robustness. At the inference stage, our model exhibits a dynamic adaptability, recalibrating the weights of models aligned with different modalities based on the saliency of each modality within the input. Contrasting with traditional multi-modal training paradigms, our approach offers a more harmonious and effective fusion of features from distinct single modalities. This synergy culminates in our method's elevated defect detection prowess.

\subsection{External Modality-guided Data Mining}

In the context of defect detection, image-based models are crucial. Recognizing the potential inaccuracies in OCR, our multimodal approach is designed to account for this. Specifically, we employ mask learning during training to simulate unstable OCR conditions, thereby enhancing the robustness of our model against OCR inaccuracies. External information, such as camera parameters and textual inscriptions, despite variability in accuracy, is integral to our comprehensive defect detection strategy. This information can contain valuable insights for defect detection, often more so than the images themselves. To leverage this, we propose an external modality-guided data mining framework, integrating additional information to assist image-based models in achieving higher detection accuracy.

\subsubsection{Modality Feature Retrieval}

The primary challenge lies in obtaining additional information from the original images, which largely depends on the data format. In our dataset and experiments, OCR is prioritized due to its simplicity, efficiency, and the ability to process information contained within the raw images without accessing other data sources.

\subsubsection{External Modality Feature Intra-Bagging} 

In this phase, our aim is to train various models based on the obtained feature modalities and the objectives of the task. We then use the features from each single modality for further feature alignment. The outputs of individual sub-models and the results of intra-bagging are taken into consideration during the decision-making and inference stages. In typical defect detection tasks, numerical features are often readily available. Performing intra-bagging by building models with similar capabilities can enhance overall performance. Gradient boosting-based decision trees, such as LightGBM \cite{lightgbm}, XGBoost \cite{xgboost}, and CatBoost \cite{catboost}, are known for their superior performance in classification or regression tasks with numerical tabular data. These trees are highly interpretable, facilitating the acquisition of critical information, with criteria closer to the root of the tree playing a significant role in decision-making and inference. When decision tree models do not exhibit stable performance, MLP is considered as an alternative. Depending on the task and data types, other models like LSTM \cite{LSTM} or Transformer \cite{transformer} may also be employed.

\begin{figure}
\includegraphics[width=\textwidth]{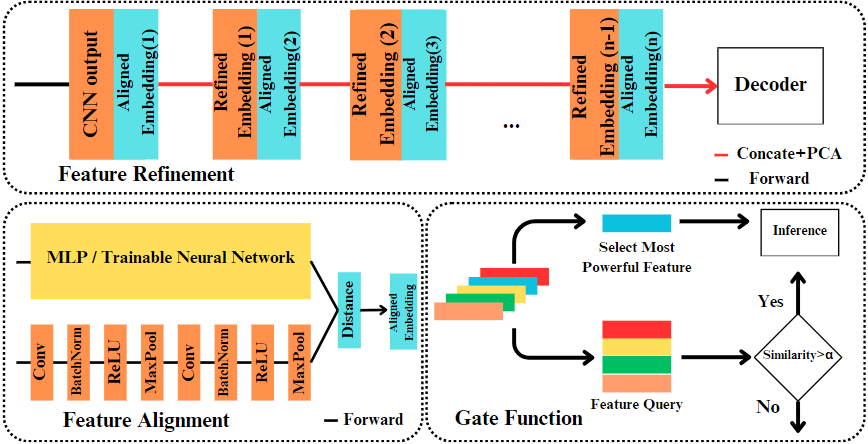}
\caption{The concepts of feature alignment, refinement, and gate function are illustrated in the following figure. External single modality data is input into the trainable encoder branch to extract features, while image features are processed in another branch. After aligning the features extracted by different modality models, they are concatenated with the output from the previous stage. Feature refinement enhances features that encompass a richer representation. Finally, the gate function is conducted to comprehensively boost overall model performance against defect detection task.} \label{feature alignment.png}
\end{figure}

\subsection{Multi-Modal Feature Fusion}
We discuss the integration and enhancement of different modalities. Integrating features from diverse modalities significantly improves zero-shot reasoning capabilities and enables pretrained models to maintain strong performance in downstream tasks. However, modality integration is a crucial issue. Several studies \cite{ALBEF,BLIP,BLIP-2} have indicated that performing feature alignment before modality integration can effectively address issues related to disparate feature distributions. By aligning and then integrating features from different modalities, we can further refine the diversity of the feature set.

\subsubsection{Image-based Encoder} 

Convolutional Neural Networks (CNNs) are the mainstream image encoders today, with popular examples including ResNet \cite{resnet}, VGG \cite{vgg}, and EfficientNet \cite{efficientnet}. The choice of an appropriate encoder is contingent upon the nature of the input data and the complexity of the task at hand. Notably, in scenarios dealing with non-defect image data and images containing defects, the patterns may not significantly differ. Furthermore, in cases of low-resolution images, selecting an encoder with excessively many parameters might result in overfitting and subsequent performance degradation on the test set.

\subsubsection{Feature Alignment} 

Features from two different modalities are encoded using corresponding feature encoders, yielding two sets of embedding vectors, as depicted in Figure \ref{feature alignment.png}. Aligning these vectors involves minimizing the distance between the two sets, thereby harmonizing their semantic representations within the feature space. The image subnetwork, for instance, utilizes several complex convolution layers due to the intricate nature of image features, while the other network adopts a simpler architecture. A distance metric function, typically based on similarity or probability-based distance, is employed for alignment; in this work, cosine similarity is used.

\subsubsection{Feature Refinement} 

Upon aligning features from multiple sub-models trained on different modalities, we enhance the model's performance by integrating information across these modalities. This involves performing PCA dimension reduction on a set of aligned vectors or embeddings encoded through image encoders, ensuring consistent lengths and concatenating them into a single vector. This process is iteratively performed multiple times to achieve feature enhancement, as illustrated in Figure \ref{feature alignment.png}.

\subsection{Inference and Decision Making Procedure}

Enhanced image features prove effective for defect detection in untested images. Nonetheless, outputs from a single modality and predictions from external modality feature intra-bagging also contribute significantly to defect detection. In certain cases, additional data may offer more crucial insights than images, but risk being underutilized or lost during training. For example, in complex scenarios requiring neural network inference for defect detection, the decision boundaries of neural networks are relatively smooth, potentially leading to false negatives. However, accessing supplementary information, such as abnormal temperatures, could allow for defect inference independent of the model. To address this, a gate function is introduced during the inference stage, adaptively adjusting the model's weights based on the obtained information, thereby ensuring high-quality defect detection.

\subsubsection{Gate Function}

The distribution of test data, which can often be unknown during the inference stage, may lead to domain shifts and poor detection performance. To mitigate this, a gate function is implemented to dynamically adjust the weights between all trained models and their outputs during inference. This function can selectively deactivate the outputs of certain models if their feature information is markedly inferior to others, ensuring balanced inference capabilities among the models. This balance is quantified by calculating the KL-divergence between the data distributions, encompassing both image data and external single or multi-modal data. Ultimately, a model ensemble is executed to bolster the overall system's robustness and achieve effective defect detection.

\section{Experiment}

\begin{figure}
\includegraphics[width=\textwidth]{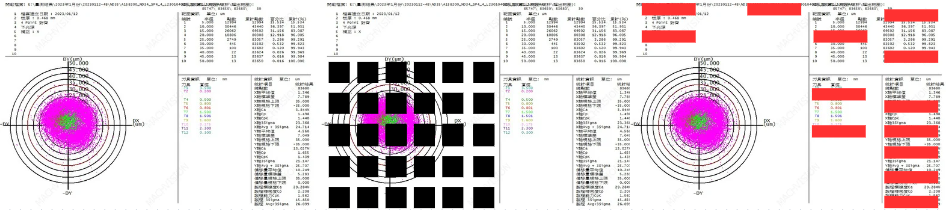}
\caption{Depicted in the left image is provided by ASE corporation. From these images, we can extract text or numerical data to gain another feature into their content. In the middle image, a grid mask is employed to introduce perturbations to areas of the image containing the target object, simulating a scenario of image corruption. The right image illustrates the  perturbations to the text portion to simulate a situation where OCR may fail.} \label{mask.png}
\end{figure}

Our dataset is provided by ASE Corporation, encompassing a total of 971 images, as shown in Figure \ref{mask.png}. These images are captured by the AOI system. Among these images, there are 587 instances without defects, while the remaining 384 images contain some defects in internal parameters or exhibit defects in the objects. The lower left corner of the provided image covers the area containing the target object. The remaining area of the image contains a wealth of parameter information about the target object.

\subsection{Implementation Details}

We divided the dataset into training and testing sets using a 7:3 ratio. During the training phase, we obtained external modal information using CnOCR \cite{CnOCR}, an efficient OCR package. We used these external modality features to train 4 defect detection models: LightGBM \cite{lightgbm}, Gradient Boosting Machines \cite{gbm}, CatBoost \cite{catboost}, and RandomForest \cite{rf}. To simplify the experimental evaluation, all hyperparameters were kept at their default values, and we also used 5-fold validation to comprehensive evaluate model's performance.

For the image-based encoder, we employ AlexNet \cite{alexnet}, VGG16 \cite{vgg}, and ResNet18 \cite{resnet}. The loss function is cross-entropy, the optimizer is SGD with a learning rate of 0.01 and weight decay of 0.0005. The models is trained for a total of 30 epochs. Our experiments were conducted using a single NVIDIA RTX TITAN.

\subsection{Performance Evaluation}

To evaluate the effectiveness of our proposed method, we conducted ablation experiments using three different image encoders, while varying the inclusion of feature alignment, refinement, and gate functions. To ensure experimental consistency and mitigate the impact of randomness, we repeated the same experiments five times and computed the average accuracy, recall, and F1-score of the prediction results as performance indicators.

The experimental results demonstrate that our proposed method significantly enhances the performance of CNN models, especially in terms of the crucial metric, recall, showing a substantial improvement, as illustrated in Table \ref{evaluation}.

\begin{table}[]
\resizebox{\columnwidth}{!}{
\begin{tabular}{c|cccccc}
\hline
\begin{tabular}[c]{@{}c@{}}Image \\ Encoder\end{tabular} &
  \begin{tabular}[c]{@{}c@{}}Feature \\ Alignment\end{tabular} &
  \begin{tabular}[c]{@{}c@{}}Feature \\ Refinement\end{tabular} &
  \begin{tabular}[c]{@{}c@{}}Gate \\Function\end{tabular} &
  Accuracy$\uparrow$ &
  Recall$\uparrow$ &
  F1-score$\uparrow$ \\ \hline
Alexnet  &    $\times$
     &     $\times$
    &     $\times$
    & 82.87 & 65.04 & 71.18 \\
Alexnet  & $\checkmark$ &    $\times$
     &  $\times$
       & 87.32 & 82.11 & 84.51 \\
Alexnet  & $\checkmark$ & $\checkmark$ &   $\times$
      &  {\color{red}90.06} & 84.55 & {\color{red}87.77} \\
Alexnet  & $\checkmark$ & $\checkmark$ & $\checkmark$ & {\color{blue}\textbf{89.38}} & \textbf{85.36} & {\color{blue}\textbf{87.13}} \\ \hline
VGG16    &     $\times$
    &      $\times$
   &      $\times$
   & 81.5  & 76.42 & 77.68 \\
VGG16    & $\checkmark$ &     $\times$
    &     $\times$
    & 86.64 & 78.04 & 83.11 \\
VGG16    & $\checkmark$ & $\checkmark$ &    $\times$
      & 83.56 & 87.8  & 81.81 \\
VGG16    & $\checkmark$ & $\checkmark$ & $\checkmark$ & \textbf{87.67} & {\color{red}\textbf{91.05}} & \textbf{86.15} \\ \hline
Resnet18 &     $\times$
    &    $\times$
     &    $\times$
     & 83.56 & 78.86 & 80.16 \\
Resnet18 & $\checkmark$ &    $\times$
     &    $\checkmark$
     & 88.01 & 83.73 & 85.47 \\
Resnet18 & $\checkmark$ & $\checkmark$ &    $\times$
     & 85.95 & 85.36 & 83.66 \\
Resnet18 & $\checkmark$ & $\checkmark$ & $\checkmark$ & \textbf{86.3}  & {\color{blue}\textbf{86.17}} & \textbf{84.12} \\ \hline
\end{tabular}
}
\caption{Performance Comparison of proposed method. Our method significantly enhances recall, which is crucial for defect detection. Feature alignment proves to be highly effective in improving accuracy, while the implementation of feature refinement along with the addition of a gate function significantly enhances recall. In the table, we marked the best and the second-best metric scores with red and blue colors, respectively.}
\label{evaluation}
\end{table}
\subsection{Robustness Analysis}
Due to our proposed method being based on varying single modality feature, it exhibits prominent performance, even in the presence of anomalies in a certain modality's data. For instance, data may become corrupted due to the failure of OCR detection  or abnormalities in the data transmission process. 

We made some perturbation on data by randomly replacing 60\% of the data detected by OCR with missing values and simulating data corruption in the acquired images using a Grid mask. Ablation experiments were performed in this context, as shown in Table \ref{robustness}. The results highlight the robustness of our method, as it remains resilient to disruptions caused by data anomalies.

\begin{table}[]
\resizebox{\columnwidth}{!}{%
\begin{tabular}{c|ccccc}
\hline
Method &
  \begin{tabular}[c]{@{}c@{}}Random Missing\end{tabular} &
  \begin{tabular}[c]{@{}c@{}}Grid Mask\end{tabular} &
  Accuracy $\uparrow$&
  Recall $\uparrow$&
  F1-Score $\uparrow$\\ \hline
VGG16    & $\times$  & $\times$  & 81.50 & 76.42 & 77.68 \\
VGG16    & $\times$  & $\checkmark$ & 75.68 & 49.59 & 63.21 \\ \hline
Our+VGG16 & $\times$  & $\times$  & 87.67 & 91.05 & 86.15 \\
Our+VGG16 & $\checkmark$ & $\times$  & 84.93 & 87.8  & 83.07 \\
Our+VGG16 & $\times$  & $\checkmark$ & 83.56 & 86.99 & 81.67 \\
Our+VGG16 & $\checkmark$ & $\checkmark$ & 82.87 & 84.55 & 80.62 \\ \hline
\end{tabular}%
}
\caption{Regarding the analysis of simulating data corruption, the experiments demonstrate that our proposed method maintains a strong performance even under extreme conditions. This demonstrates its robustness and ability to deliver consistent results in challenging scenarios.}
\label{robustness}
\end{table}
\section{Conclusion}

In this work, we have presented OANet (OCR-AOI-Net), a framework that leverages a single-modality-data-aware multimodal approach. At the heart of our method is the use of Optical Character Recognition (OCR) to extract features from various modalities, enabling the capture of a broad spectrum of semantic information from raw images. This comprehensive approach ensures that vital information is not overlooked, leading to more reliable and consistent predictions in defect detection tasks. Empirical evaluations highlight the efficacy of our method, particularly in enhancing the recall metric. Notably, our model demonstrates resilience, maintaining robust performance even in scenarios where some modality data is compromised. We believe that further exploration and integration of external multi-modal features will significantly enhance defect detection capabilities, particularly in reducing the incidence of false negatives in detection procedures.


%
%
%
%

\end{document}